# Problem-Focused Incremental Elicitation of Multi-Attribute Utility Models

**Vu Ha and Peter Haddawy**
Decision Systems and Artificial Intelligence Lab[*]
Dept. of EE&CS
University of Wisconsin-Milwaukee
Milwaukee, WI 53201
{*vu,haddawy*}*@cs.uwm.edu*

## Abstract

Decision theory has become widely accepted in the AI community as a useful framework for planning and decision making. Applying the framework typically requires elicitation of some form of probability and utility information. While much work in AI has focused on providing representations and tools for elicitation of probabilities, relatively little work has addressed the elicitation of utility models. This imbalance is not particularly justified considering that probability models are relatively stable across problem instances, while utility models may be different for each instance. Spending large amounts of time on elicitation can be undesirable for interactive systems used in low-stakes decision making and in time-critical decision making. In this paper we investigate the issues of reasoning with incomplete utility models. We identify patterns of problem instances where plans can be proved to be suboptimal if the (unknown) utility function satisfies certain conditions. We present an approach to planning and decision making that performs the utility elicitation incrementally and in a way that is informed by the domain model.

## 1   INTRODUCTION

Decision-theoretic problem solving requires a probabilistic model of the world and of actions and a utility model specifying the objectives to be achieved[1]. While the probability model is relatively stable across problem instances, the utility model for each problem instance specifies the current objectives, and thus must be elicited anew. While much work has been done within the UAI community to address the problem of eliciting probabilities [14, 4], somewhat paradoxically little has been done to address the elicitation of utility models. While this problem has been extensively studied in the field of Decision Analysis [8, 12], eliciting utility functions for interactive decision making systems raises new issues and offers new ways to facilitate elicitation, as we demonstrate in this paper.

---

[*]This work was performed while the authors were on leave at the Dept. of Applied Statistics, National Institute of Development Administration, Bangkok, Thailand.

[1]Note that these models need not be numeric.

Elicitation of utility models can be a tedious and time-consuming task. Most planning and decision making systems that make use of utilities nevertheless assume that all the utility information the system will receive is provided up front, thus all elicitation must be performed before the system provides the user with any information concerning the problem solution. Indeed, most systems assume that a *complete* specification of the utility or value function is provided before any reasoning can proceed [13, 3]. This paradigm has the following drawbacks:

- Often partial utility information will be sufficient to narrow down the set of alternatives to the point that the user is sufficiently indifferent among the remaining options to simply choose one or to the point that the user can easily pick out his most preferred option. For example, consider a vacation planning assistant that helps a user to find a suitable vacation package. To escape the icy Wisconsin winter I may wish to go to some place warm and tropical with good diving. I may also wish to minimize my travel time and the cost of the trip. Clearly I would like to minimize the amount of time I need to spend specifying my preferences to the system. Fortunately, the above criteria are relatively easy to specify and can be used to eliminate a large number of undesirable options, without requiring the user to specify tradeoff information such as the tradeoff between travel time and quality of diving. Some tradeoff information may be necessary to sufficiently narrow down the set of options, but we maintain that this can be best identified by first filtering the options through the partial preference model.

- In decision making situations where timeliness is of critical importance (for example, military

battlefield planning), we may not have sufficient time to completely elicit a utility function before presenting the user with some form of solution. While researchers have addressed the problem of time-critical decision making by designing flexible algorithms for reasoning under time constraints [7, 5], for many problems the elicitation of the model takes far more time than the actual inference. In such a situation we would like a flexible method for eliciting utility information that permits us to take advantage of whatever time is available and that elicits the most useful information first.

The ability to reason with partial utility models thus seems to be of no little importance for decision analysis and decision-theoretic planning systems. Recent work on qualitative represention of preferences [11, 1, 2] has addressed this issue by focusing on providing formal languages for representing partial preference models. In this paper we take a more modest approach to the representation of partial preference models in order to present a complete solution to elicitation and decision making. We work within the standard framework of multi-attribute utility theory and assume that the subutility functions for the relevant outcome attributes are known. First we identify various problem characteristics that permit suboptimality of plans to be inferred based on knowledge of the subutility functions. Then we examine how additional tradeoff information to further eliminate suboptimal plans can be elicited incrementally.

The problem of eliminating suboptimal plans by using only knowledge of the subutility functions has been addressed by work on the DRIPS planner [6]. Assuming a simple additive utility function, DRIPS iteratively cycles through the process of refining abstract plans and eliminating classes of dominated plans until only the efficient frontier remains. The main drawback of this system, however, is that it does not allow the partial utility model to be augmented, and as such is incapable of narrowing down the efficient frontier when the need for more specificity arises. The second aim of this paper is to address this problem. In particular, we present a *lazy elicitation* approach to planning and decision making that alternates between eliminating provably suboptimal plans and interactively acquiring information about the utility model from the user. The process is terminated when the user indicates that the current set of candidate plans is acceptable. A novelty of this approach is that the decision of what piece of information about the utility model should be acquired is aided with the knowledge about the still-competing alternatives - the efficient frontier. More specifically, we propose applying the *rank correlation coefficient* measure - a well-known concept in regression theory - to identify the piece of tradeoff information that would reveal to us the highest number of suboptimal plans. We present an intuitive justification for the use of this measure and support it with empirical results. We have extended the DRIPS planner to incrementally acquire tradeoff information using this measure. We demonstrate its effectiveness on the medical decision problem of finding optimal strategies for management of deep venous thrombosis.

## 2    PARTIAL UTILITY MODELS

In this section we investigate the issues of reasoning with incomplete utility models. With a complete utility function, we can determine our preference between any two candidate plans by simply computing and comparing their expected utilities. The plan with lower expected utility is deemed dominated and is eliminated. We call this *overall dominance*. But when only a partial specification of the utility function is available, what we might be able to determine is some *local dominances*. For example, we may think that the plan of going to Siberia for winter vacation is dominated by the plan of going to the Caribbean with respect to individual objectives "warmth" and "distance". So the question now becomes: "what sorts of conditions must be satisfied in order for us to legitimately infer overall dominance based on a set of local dominances?".

In multi-attribute utility theory, "local dominance", or local preference, implicitly presupposes certain kinds of independencies. We continue with a quick review of fundamental concepts and terminology of multi-attribute utility theory. For an extensive developement of this theory, the reader is refered to [8]

### 2.1    MULTI-ATTRIBUTE UTILITY THEORY

In the most abstract form, multi-attribute utility theory is concerned with the evaluation of the consequences or outcomes of an agent's decisions or acts, where outcomes are characterized with complex sets of features called *attributes*. The attributes are henceforth denoted by $X_1, X_2, \ldots, X_n$, and outcomes are designated by $x = (x_1, \ldots, x_n)$, where $x_i$ designates the value of attribute $X_i$. Abusing notation, we denote the set of values an attribute $X_i$ can take simply by $X_i$, and thus the outcome space $\Omega$ is just the Carthesian product $X_1 \times X_2 \times \cdots \times X_n$. We will often talk about subsets $Y$ of the set of attributes $X = \{X_1, X_2, \ldots, X_n\}$, and also refer to $Y$ and their complements $Z = X - Y$ as attributes. With respect to such a pair $(Y, Z)$, an outcome $x = (x_1, x_2, \ldots, x_n)$ can be written as $(y, z)$. For example, if $n = 5$ and $Y = \{X_1, X_3\}$, then $y = (x_1, x_3)$ and $z = (x_2, x_4, x_5)$.

When there is no uncertainty involved, in order to make decisions, an agent needs only express preferences among outcomes. The preference relation, henceforth denoted by $\succeq$, can often be captured by an order-preserving, real-valued *value function* $v$.

Given a preference order $\succeq$ over $\Omega$, an attribute $Y \subset X$

is *preferentially independent* of its complement $Z$, or, for short, $Y$ is PI, if the preference $\succeq$ over outcomes that are fixed in $Z$ at some level does not depend on this level.

To infer the overall preference from such local preferences also seems straightforward; what we need is the direction agreement of the local preferences. Formally, we have the following proposition.

**Proposition 1** *If both $Y$ and its complement $Z$ are PI, and $y' \succeq y'', z' \succeq z''$, then $(y', z') \succeq (y'', z'')$.*

When there is uncertainty, the outcomes of an agent's decisions or acts are characterized by probababilities. To differentiate between certain and uncertain outcomes, we call uncertain outcomes *prospects*, and use *outcomes* to refer to outcomes with no uncertainty. Thus, prospects are probability distributions over possible outcomes, or outcome space $\Omega$. Now the agent faces the more difficult task of ranking the prospects, instead of outcomes. The central result of utility theory is a representation theorem that proves the existence of a *utility function* $u : \Omega \to R$ such that preference order among prospects can be established based on the expectation of $u$ over outcomes. The key point here is that the utility function is defined over outcomes alone; the extension to prospects via expectation is a consequence of the axioms of probability and utility [10].

The generalization of preferential independence to the uncertainty case is the concept of *utility independence* (UI). Given a preference order $\succeq$ over the prospects over $\Omega$, an attribute $Y \subset X$ is *utility independent* of its complement $Z$, or, for short, $Y$ is UI, if the preference $\succeq$ over prospects that are fixed in $Z$ at some level does not depend on this level.

When applicable, utility independence gives us very useful hints about the form of the utility function. Towards this end, below we list a few relevant results, using [8] as our source.

**Proposition 2 (Basic Decomposition)** *If some attribute $Y \subset X$ is UI, then the utility function $u(x)$ must have the form:*

$$u(x) = u(y, z) = g(z) + h(z).u(y, z^+),\,^2$$

*where $g(\cdot)$ and $h(\cdot) > 0$ depends only on $z$ but not on $y$, and $z^+$ is some fixed value of $z$.*

From this result and the next few ones, we can see that utility independence plays just as fundamental a role in utility theory as does probabilistic independence in probability theory: it provides modularity and decomposition. When the UI of an attribute $Y$ holds, the assessment of the utility function $u(x)$ is reduced to the assessment of three functions all of which have fewer arguments. This reduction of dimensionality is crucial, both analytically and practically. The following theorems further investigate the implications various sets of UI assertions have on the form of the utility function.

---

²The function $u(y, z^+)$, defined over variable $y$ is called the *subutility function* for the attribute $Y$, and is sometimes designated by $u_Y(y)$.

**Proposition 3 (Multi-Linear Form)** *If each attribute $X_i$ is UI, then $u(x)$ can be written in the following form:*

$$u(x) = \sum_{\emptyset \neq Y \subseteq X} k_Y . \prod_{X_i \in Y} u_i(x_i), \qquad (1)$$

*where $u_i(x_i)$ is the utility function $u(x_i, \overline{x_i}^+)$, defined at some fixed, convenient value $\overline{x_i}^+ \in \overline{X_i} = X - \{x_i\}$.*

The $k_Y$ constants serve as scaling constants so that all subutility functions $u_i$ are scaled from 0 to 1 wrt any 2 values $x_i^0$ and $x_i^*$ so that $x_i^*$ is prefered to $x_i^0$, and $u$ is scaled from 0 to 1 wrt $(x_1^0, \ldots, x_n^0)$ and $(x_1^*, \ldots, x_n^*)$. In practice, it is often convenient to set $x_i^*$ and $x_i^0$ to the two extreme levels of $X_i$.

One problem of the multilinear utility function is that it requires the assesment of $O(2^n)$ scaling constants, in addition to the assesment of the subutility functions $u_i$. Stronger independence assumptions must hold in order for simpler forms of utility functions to be valid. One of the most interesting cases is *mutual utility independence* (MUI), when not only each single attribute $X_i$ is UI, but every attribute $\emptyset \neq Y \subset X$ is UI.

**Proposition 4 (MUI Forms)** *If the attributes $X_i$ are MUI, then there are $n+1$ constants $k, k_1, k_2, \ldots, k_n$ such that $1 + k = \prod_{i=1}^{n}(1 + kk_i)$, and $k_Y = k^{|Y|-1} \prod_{x_i \in Y} k_i$. And, depending on the value of $k$, the utility function takes one of the following two forms:*

- *If $k \neq 0$, then $1 + ku(x) = \prod_{i=1}^{n}(1 + kk_iu_i(x_i))$, and the utility function is said to have* multiplicative form.

- *If $k = 0$, then $u(x) = \sum_{i=1}^{n} k_iu_i(x_i)$, $k_i \geq 0$, $\sum_{i=1}^{n} k_i = 1$, and the utility function is said to have* additive form.

## 2.2 REASONING WITH PARTIAL UTILITY MODELS

We now focus our attention on the issue of how to infer overall dominance from a set of local dominances. From now on we assume that each attribute $X_i$ is UI. In other words, the utility function is multilinear (see Equation 1). We assume that $pl_1$ and $pl_2$ are two candidate plans that result in the joint distributions over $\Omega$ with the corresponding density functions $f_1$ and $f_2$. Naturally, we would like to look for an analogous version of Proposition 1. We begin with the following definition.

**Definition 1 (Local Dominance)** *Plan $pl_1$ is said to dominate plan $pl_2$ with respect to attribute $X_i$, denoted $pl_1 \succeq_i pl_2$, if*

$$\int_\Omega u_i(x_i)f_1(x)dx \geq \int_\Omega u_i(x_i)f_2(x)dx.$$

This means that if $u_i(x_i)$ were the overall utility function, then $pl_1$ would dominate $pl_2$. This kind of local dominance can easily be established when the subutility functions $u_i$ are known but the scaling constants $k$ are yet to be determined, a situation that often occurs in the process of assessing the overall utility function $u$. A natural step is to see if we can infer overall dominance from a set of such local dominances, without the knowledge of the scaling constants. The next proposition answers this question in the affirmative, provided that the utility function has additive form.

**Proposition 5** *Suppose that $pl_1 \succeq_i pl_2, \forall i = 1, \ldots, n$. Furthermore, suppose that the utility function $u$ has additive form. Then $pl_1 \succeq pl_2$.*

*Proof:* For $j = 1, 2$ we have:

$$
\begin{aligned}
E[u(pl_j)] &= \int_\Omega u(x)f_j(x)dx \\
&= \int_\Omega (\sum_{i=1}^n k_i u_i(x_i))f_j(x)dx \\
&= \sum_{i=1}^n k_i \left( \int_\Omega u_i(x_i)f_j(x)dx \right)
\end{aligned}
$$

Thus $pl_1 \succeq pl_2$ since the coefficients $k_i$ are non-negative. $\square$

In the case when the utility function is only multi-linear, an extra condition needs to hold in order to infer overall dominance.

**Proposition 6** *Suppose that $pl_1 \succeq_i pl_2, i = 1, \ldots, n$. Furthermore suppose that in the joint distributions $f_1$ and $f_2$, the random variables $x_i$ are probabilistically independent. Then $pl_1 \succeq pl_2$.*

*Proof:* Note that the multi-linear function $u$ in Equation 1 can be considered as the composition $u = h \circ w$, where $w : \Omega \rightarrow R^n$, $w(x) = (u_1(x_1), \ldots, u_n(x_n))$, and $h : R^n \rightarrow R$, $h(w_1, w_2, \ldots, w_n) = \sum_{\emptyset \neq Y \subset X} k_Y \prod_{X_i \in Y} w_i$. From the Basic Decomposition (Proposition 2), and the well-known fact that utility functions are unique up to a positive linear transformation, it is not hard to see that $h$ is a monotonically non-decreasing function with respect to *each* of its variables.

With the introduction of this decomposition, the expected utility of plan $pl_j, j = 1, 2$ can be written as:

$$E[u(pl_j)] = E_{f_j}[h \circ w(x)]$$

$$
\begin{aligned}
&= \sum_{\emptyset \neq Y \subset X} k_Y E_{f_j}[\prod_{X_i \in Y} u_i(x_i)] \\
&\quad \text{(because $x_i$ are prob. indep. wrt $f_j$)} \\
&= \sum_{\emptyset \neq Y \subset X} k_Y \prod_{X_i \in Y} E_{f_j}[u_i(x_i)] \\
&= h(E_{f_j}[u_1(x_1)], \ldots, E_{f_j}[u_n(x_n)]).
\end{aligned}
$$

The overall dominance of $pl_1$ over $pl_2$ then follows directly from the given local dominances $\succeq_i$, and the monotonicity of the $n$-dimension function $h$. $\square$

It is interesting to note the intrinsic relationship between Proposition 5 and 6. To see this, we first note that a necessary and sufficient condition for the utility function $u$ to have additive form is that the preference among prospects over $\Omega$ depends only on their marginal distributions [8]. We then note that for a joint distribution where the individual random variables are probabilistically independent, the joint distribution will be completely determined based on its marginals. The key difference between the two propositions is that we impose different restrictions in their premises. In the former proposition, the restriction is imposed on the form of the *utility function* (it must be additive), while in the latter, the restriction is imposed on the *plans* (they must result in distributions where probabilistic independence holds). So in a sense, the two results are complementary.

The above two propositions help us to infer overall dominance from local dominance. But they seem to require rather stringent conditions, either on the utility function or the plans. If the utility function has multiplicative form, i.e. when we assume only MUI, unfortunately it is possible that $pl_1$ dominates $pl_2$ with respect to each individual attribute, yet $pl_2$ dominates $pl_1$ overall. Consider the following example. Let $X = \{Y, Z\}$, $\Omega = \{(y_1, z_1), (y_2, z_2), (y_3, z_3)\}$. The overall utility function $u$ is the product of the subutility functions, $u_Y$ and $u_Z$: $u(y, z) = u_Y(y)u_Z(z)$ [3]. The utility functions and the two density functions $f_1$ and $f_2$ are specified in the following table [4].

| Y | Z | $u_Y$ | $u_Z$ | u | $f_1$ | $f_2$ |
|---|---|---|---|---|---|---|
| $y_1$ | $z_1$ | 1 | 0 | 0 | 1/2 | 1/4 |
| $y_2$ | $z_2$ | 0 | 1 | 0 | 1/2 | 1/4 |
| $y_3$ | $z_3$ | 1/3 | 1/3 | 1/9 | 0 | 1/2 |

Clearly, $f_1$ dominates $f_2$ with respect to both $Y$ and $Z$ ($E_{f_1}[u_Y] = E_{f_1}[u_Z] = 1/2 > E_{f_2}[u_Y] = E_{f_2}[u_Z] = 5/12$), but overall $f_2$ dominates $f_1$ ($E_{f_1}[u] = 0 < E_{f_2}[u] = 1/18$). It is interesting to ask if there is a condition that would ensure the inference, yet is weaker than probabilistic independence.

---

[3] This can be derived from the multiplicative form, whenever $k$ is positive.

[4] We also have a counter example for the case when the functions are continuous.

To summarize so far, we have proved that overall dominance can be infered from local dominances if the utility function has additive form, or the random variables corresponding to the attributes are probabilistically independent.

# 3   PROBLEM-FOCUSED INCREMENTAL UTILITY ASSESMENT

In this section we address the issue of acquiring more information about the utility model in order to narrow down the efficient frontier. We focus our attention on the case when the utility function has additive form: $u(x) = \sum_{i=1}^{n} k_i u_i(x_i)$, $k_i \geq 0$, $\sum_{i=1}^{n} k_i = 1$. We also assume that the subutility functions $u_i$ are already assessed.

Recall that for a plan $pl$ that results in a prospect that is a probability distribution with density function $f$, its expected utility can be written as: $E[u(pl)] = \sum_{i=1}^{n} k_i w_i$, where $w_i = \int_{\Omega} u_i(x_i) f(x) dx$. We set $\vec{k} = (k_1, k_2, \ldots, k_n)$ and $\vec{w} = (w_1, w_2, \ldots, w_n)$, and thus $E[u(pl)] = \vec{k} \circ \vec{v}$, i.e., the inner product of two vectors $\vec{k}$ and $\vec{w}$.

Let us now assume that we have to choose the optimal plan(s) from a set $PL = \{pl_1, pl_2, \ldots, pl_m\}$, where each plan $pl_j$ has the expected utility $E[u(pl_j)] = \vec{k} \circ \vec{w}_j$. Using Proposition 5, we can infer overall dominance from local dominances, i.e., if $pl_i, pl_j \in PL$ such that $\vec{w}_i$ dominates $\vec{w}_j$ with respect to each component, then we can conclude that overall, $pl_i$ dominates $pl_j$. We can also view this decision making problem as one with certainty. The objective is characterized with $n$ attributes $W_1, W_2, \ldots, W_n$, where a plan $pl$ results in the (certain) outcome $\vec{w}$, and there is an additive value function $v(\vec{w}) = \vec{k} \circ \vec{w}$ [5].

## 3.1   THE RANK CORRELATION COEFFICIENT

When the efficient frontier identified by the above method is still too large, there is clearly a need to acquire more information about the utility function $u$ (in this case, information about the scaling constants $k_i$) in order to eliminate more plans. In practice, there are two types of questions that we use to ask the decision maker.

- **Question Type I.** For what probability $p$ are you indifferent between:
    1. the prospect that has $p$ chance at $x^*$ and $1-p$ chance at $x^0$ and
    2. the outcome $(x_i^0, \overline{x_i}^*)$.

[5]Using Proposition 1, we have another proof for Proposition 5.

- **Question Type II.** Suppose that $x_i' \in X_i$. For what value $x_j' \in X_j$ are you indifferent between two outcomes that yield $x_i', x_j^0$, and $x_i^0, x_j'$ respectively, and that agree on all other attributes at some level [6].

It can be shown that the answer to Question I will be equal to $k_i$, and since it can be shown that $k_i u_i(x_i') = k_j u_j(x_j')$, the answer to Question II would reveal the ratio $k_i/k_j$ ($= u_j(x_j')/u_i(x_i')$).

For our purpose, which is to identify more suboptimal plans, the value of a single scaling constant $k_i$ does not seem to be of much help. However, with the knowledge of the *tradeoff ratio* $k_i/k_j$ for some two attributes $X_i$ and $X_j$, the expected utility of a plan $pl$ can be rewritten as:

$$E[u(pl)] = \sum_{l=1, l \neq i, j}^{n} k_l w_l + k_j(\frac{k_i}{k_j} w_i + w_j) = \vec{k}' \circ \vec{w}',$$

where $\vec{k}'$ is obtained from $\vec{k}$ by deleting the $i$th component, and $\vec{w}'$ is obtained from $\vec{w}$ by deleting the $i$th component and changing the $j$th component to $\frac{k_i}{k_j} w_i + w_j$. Note that the modified vectors will now have $(n-1)$ components. We call this step the *merging of two attributes $X_i$ and $X_j$*. The problem now reduces to a decision making problem with certainty, and $n-1$ attributes. The value function for this problem has additive form, with unknown scaling constants. In other words, we are dealing with exactly the same type of problem, with one less dimension. The process of identifying the new efficient frontier can now be resumed. The process of alternatively finding the efficient frontier and attribute-merging can be repeated until the user feels that the set of candidate plans is small enough, or until all of the scaling constants (and thus the overall utility function) are completely specified.

Another, perhaps more illustrative way to look at this problem is the following. We view each plan as a competitor in some competition and each attribute as a member of the jury. Initially, there are $m$ competitors and $n$ jurors. The expected utility value $v_{ji}$ that the plan $pl_j$ gets for the attribute $X_i$ will now be interpreted as the mark competitor number $j$ gets from juror number $i$. The marks of the jurors, however, are not of equal importance but instead are weighted by unknown weights $k_i$. The attribute-merging step requires the acquisition of information, specifically the ratio of some two weights $k_i, k_j$. We can view this step as the unification of the corresponding two jurors, number $i$ and $j$ to a group with the responsibility of giving a single mark to each competitor. If our goal is to find a unification that reveals a high number of non-contender competitors, then obviously, we would like to avoid choosing a pair of jurors whose rankings for competitors - according to their marks - highly agree.

[6]Note that Question II is not applicable in the case when the two attributes $x_i$ and $x_j$ are discrete-valued. If such situation occurs, we need to use Question Type I to assess the coefficients for the two attributes and then compute their ratio.

So what we need here is a measure of the degree to which two rankings (dis)agree. Intuitively, one possible measure is the number of pairs of competitors who get "reverse" rankings from the two jurors. For example, if the first juror ranks four plans in order $(1, 2, 3, 4)$, and the second juror ranks the same four plans in order $(1, 3, 4, 2)$, then there are two "reverse" pairs, $(3, 2)$ and $(4, 2)$. The more "reverse" pairs there are, the more the two jurors disagree. In regression theory, the *rank correlation coefficient* (RCC), defined below, is an easily computed measure of this [7].

**Definition 2 (Rank Correlation Coefficient)**
*Suppose that $a = (a_1, a_2, \ldots, a_n)$ and $b = (b_1, b_2, \ldots, b_n)$ are two permutations of $\{1, 2, \ldots, n\}$. Then the rank correlation coefficient of $a$ and $b$, denoted by $\rho(a, b)$ is defined as:*

$$\rho(a, b) = 1 - \frac{6 \sum_{i=1}^{n} (a_i - b_i)^2}{n^3 - n}$$

It is not hard to see that $-1 \leq \rho(a, b) \leq 1$. In fact, the more "similar" the two rankings $a$ and $b$ are (the smaller the sum $\sum (a_i - b_i)^2$ is), the greater the rank correlation coefficient is. When there is no disagreement, i.e. when $a_i = b_i, \forall i$, the coefficient attains the highest possible value, 1. The coefficient attains the lowest possible value, -1, when $a_i + b_i = n + 1, \forall i$, i.e. when the two rankings are reverse of one another.

**Example.** Consider the following three rankings: $a = (1, 2, 3, 4)$, $b = (1, 3, 4, 2)$, and $c = (4, 2, 1, 3)$. We have that $\rho(a, b) = 0.4 > \rho(a, c) = -0.4$. It means that the agreement between $a$ and $b$ is higher than the agreement between $a$ and $c$. This is no suprise because with respect to $a$, $b$ contains two reverse pairs $((3, 2)$ and $(4, 2))$, while $c$ contains four reverse pairs $((4, 1), (4, 2), (4, 3), (2, 1))$

Thus, by computing the RCC of every possible pair of attributes, we can identify the two attributes that when merged would likely reveal the most suboptimal plans. The process of alternatively asking for the tradeoff ratio of two attributes that have the smallest RCC, and identifying & eliminating suboptimal plans is repeated until the user is satisfied with the current set of candidates.

It is important to note that the use of RCC does not guarantee that we will be able to identify the most suboptimal plans; it is only a heuristic. Although the merging of two highly conflicting attributes will resolve their conflict, the unpredictability of the tradeoff ratio $(k_i/k_j$, which in principle can be any positive number) makes it possible that a different pairing would eliminate more plans. For example, one attribute may conflict highly with all other attributes but the single attribute that conflicts most with it may have a very small coefficient so that the ranking of the merged attribute is still very close to that of the first. However, we are encouraged by the experimental results of using RCC, reported in the next subsection.

A point worth mentioning is that there are situations where the decision maker may feel more comfortable or confident in estimating the tradeoff ratios among certain types of attributes. For example, he may be comfortable in tradeoffs that involves monetary values such as costs while reluctant and uncertain in trade-offs that involves less straightforward or sensitive attributes such as morbidity or mortality. In such cases, the RCC method serves as a hint instead of as a normative must.

## 3.2 EXPERIMENTAL RESULTS

In our first experiment, we compare the performance of the algorithm that uses RCC with the random algorithm (RAND), which randomly picks two attributes to merge. The performance of the algorithms is measured by the number of plans they are able to eliminate after merging the first pair of attributes. We take the performance of the omniscient algorithm (OPT), which knows exactly the best two attributes to merge, as the baseline.

Each experiment instance is generated by first choosing $n$, the number of attributes and $m$, the size of the initial efficient frontier. We examine the cases when $n$ takes values from 4 to 8 and $m$ takes values from $\{25, 50, 100, 200\}$ since these are ranges of values we expect to see in real-world problems. For each such pair of $(m, n)$ we randomly generate - using a uniform distribution on the interval $[0, 1]$ - $n$ coefficients $k_i$ and $m \times n$ expected subutilities for the $m$ plans according to each of the $n$ attributes.

One standard way to evaluate the performance of two algorithms is to compute their *competitive ratios*[8]. We ran 500 trials and computed the average ratios of the number of suboptimal plans RCC and RAND identified versus the number OPT identified: $\rho(RCC) = .89$ and $\rho(RAND) = .65$. Other informative statistics are the percentage of times when RCC outperforms RAND: 85%, the percentage of times RCC identifies as many suboptimal plans as OPT does: 37%, and the percentage of times RCC does better than average: 92%. Here average is defined as the average number of plans eliminated over all possible choices of attribute pairings.

---

[7]It can be shown that the more reverse pairs there are, the smaller the rank correlation coefficient is.

[8]Competitive ratio is the fundamental concept in the competitive analysis of on-line algorithms. This ratio measures the performance of an on-line algorithm using that of the optimal off-line algorithm as the straw man. In this context, OPT plays the role of the optimal off-line algorithm.

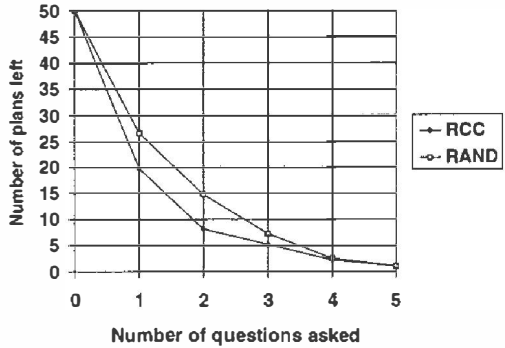

Figure 1: Anytime performance of RCC and RAND.

In our second experiment, we compare the performances of RCC and RAND when the process of eliminating suboptimal plans and merging attributes is repeated until all of the attributes are merged into a single comprehensive attribute. Figure 1 records the results in the case when $m = 50$ and $n = 6$. The two graphs indicate the average sizes - also with 500 problem instances - of the current efficient frontier after 0,1,2,3,4, and 5 attribute-mergings are done. The fact that the graph of RCC lies below the graph of RAND indicates that RCC on average has better anytime performance than RAND: using a same number of questions, RCC is on average able to identify more suboptimal plans than RAND. This is also true for *any* other case of $\{m, n\}$ pairs.

We have enhanced the DRIPS planning system, incorporating the ability to incrementally add tradeoff ratios of attributes during the process of eliminating suboptimal plans, using RCC as its guiding heuristic. We applied this new version of DRIPS to the problem of identifying the optimal management strategy for deep venous thrombosis [6]. Strategies for this problem are evaluated with respect to four attributes: mortality (DEATH), progression to pulonary embolism (PE), incidence of a bleeding episode (BLEED), and monetary cost of test and treatment (COST). PE and BLEED can be considered morbidity factors. The first three attributes are binary attributes, while COST has a range from 0 to 50,000.

The overall utility function for the strategies is assumed to have additive form, with the subutility functions for the attributes: $u(attr) = 1 - attr$ for the binary attributes and $u(COST) = 1 - COST/50,000$. Out of 1022 initial plans, the system identifies the initial efficient frontier containing 91 candidate plans. The system then identifies BLEED and PE as the most conflicting pair of attributes. Since these two attributes are discrete, we need to assess the coefficients for each of the two attributes using Questio Type I. The coefficient for attribute BLEED, for example, can be assessed by asking the decision maker the following question:

"For what probability $p$ are you indifferent between a lottery that yields either the outcome $\langle DEATH = 0, BLEED = 0, PE = 0, COST = 0\rangle$ with probability $p$ and outcome $\langle DEATH = 1, BLEED = 1, PE = 1, COST = 50,000\rangle$ with probability $1 - p$, and the certain outcome $\langle DEATH = 1, BLEED = 0, PE = 1, COST = 50,000\rangle$?"

The answer for this question will be the coefficient of the attribute BLEED.

After getting the coefficients for BLEED (.01) and PE (.02), we can compute the tradeoff ratio between these two attributes (1/2). Using this ratio, the system is able to eliminate 66 more plans, and thus reduces the set of candidates to 25 plans. After getting the second tradeoff ratio of attributes BLEED/PE and DEATH, the system narrows down the set of candidates to 4 plans [9]. This experiment demonstrates the ability of the DRIPS system to quickly identify a set of candidate plans with a very small size without obtaining the complete utility function (namely, without obtaining the tradeoff information that involves attribute COST).

## 4    SUMMARY AND RELATED WORK

In this paper we have explored possible ways to infer overall dominance from a set of local dominances using only partial utility models. We introduce the concept of lazy, problem-focused utility elicitation and show how this concept is used in elicitation for decision making problems where the utility function has additive form. We propose using the rank correlation coefficient to identify the piece of information that would most likely reveal a large number of suboptimal candidates. We demonstrate its effectiveness with experimental results.

It is interesting to note that in decision making problems under certainty where the value function has additive form, the RCC method is particularly useful. In order to identify the efficient frontier and determine the rank correlation coefficient, we need not know the exact forms of the subvalue functions; all we need is the directionalities of the subpreferences. The subvalue functions need to be assessed only when we ask the decision maker for the tradeoff ratio between the two chosen attributes. As a consequence, it is quite possible that the interactive planning process can be terminated without assessing all the subvalue functions - a clear savings in effort.

Additive utility functions have enormous computational advantages over other less structured utility functions such as multilinear or multiplicative utility functions. However, the condition for the existence of an additive utility function, additive independence, is rather strict. A possible bridge between additive utility functions and the rest is the concept of *conditional additive independence* (CAI), as introduced in [8], and recently further explored in [1]. Conditional additive independence is weaker than additive independence, but still provides additive decompositions for utility functions, albeit in a slightly different form. A straightforward extension of this work would be to investigate the extension of the RCC method to the case of CAI.

---

[9] In this case, since we have enough data to determine *all* of the coefficients, we have the complete utility function and thus are able to identify the optimal plan(s).

Linden, et.al. [9] present a general methodology and a particular implementation of interactive problem solving very much in the spirit of the present work. Their candidate/critique model works by presenting a set of candidate solutions to the user and then incrementally eliciting user preferences in the form of critiques of these solutions. They assume an additive utility function and start with some user preferences augmented with a set of default preferences. In addition to displaying the option that is optimal with respect to these preferences, they use two heuristics to select other interesting options. The first is to display options that are significantly different from one another with respect to the utility function. The second is to display extreme solutions that optimize at least one attribute. For example, in their domain of flight scheduling they always display the cheapest flight. Use of these heuristics provides the user with a set of candidate solutions that provide a fair coverage of the space of possible solutions. The user provides critiques of the solutions by directly modifying the represented preferences using a graphical interface.

The idea of presenting the user with extreme solutions could be nicely incorporated into our present work. In particular, it may be that the ranges of some attributes are so small that they can be ignored for all practical purposes. For example, if we are considering airline flights and the range of cost for the available options is only $500 – $510, then cost should not be a factor. By displaying the possible ranges of outcome attributes, we can permit the user to identify those for which there is no significant difference among solutions and can set it's weight to zero. The same effect could be accomplished by eliciting tradeoff information, but setting the weight to zero involves a simpler decision on the part of the user.

### Acknowledgements

We would like to thank AnHai Doan, Sumanta Guha, and Ethan Munson for helpful discussions. This work was partially supported by a Fulbright fellowship to Haddawy, by a grant from Rockwell International Corporation, and by NSF grant IRI-9509165.